\documentclass[11pt,onecolumn,a4paper]{article}

\usepackage{amsmath}
\usepackage{amsfonts}
\usepackage{amssymb}
\usepackage{amsthm}

\usepackage[ansinew]{inputenc}
\usepackage{graphicx,color}
\usepackage{url}
\usepackage{latexsym,enumerate}
\usepackage{booktabs}
\usepackage{framed}
\usepackage{subfigure}
\usepackage{multirow} 
\usepackage{color}
\usepackage{colortbl}
\definecolor{gray1}{rgb}{0.8,0.8,0.8}
\definecolor{gray2}{rgb}{0.95,0.95,0.95}
\newcommand\degree{}
\renewcommand{\degree}{\ensuremath{^\circ}}

\begin{document}

\title{Separating the Real from the Synthetic: Minutiae Histograms as Fingerprints of Fingerprints}

\author{Carsten Gottschlich\thanks{Institute for Mathematical Stochastics,
University of G\"ottingen,
Goldschmidtstr. 7, 37077 G\"ottingen, Germany.
Email: gottschlich@math.uni-goettingen.de} \ and Stephan Huckemann\thanks{Felix-Bernstein-Institute 
for Mathematical Statistics in the Biosciences,
University of G\"ottingen,
Goldschmidtstr. 7, 37077 G\"ottingen, Germany.
Email: huckeman@math.uni-goettingen.de}}

\date{}

\maketitle


\begin{abstract}

In this study\footnote{This is a preprint version of \cite{GottschlichHuckemann2015}.} we show that by the current state-of-the-art synthetically generated fingerprints 
can easily be discriminated from real fingerprints.
We propose a non-parametric distribution based method using second order extended minutiae histograms (MHs)
which can distinguish between real and synthetic prints with very high accuracy.
MHs provide a fixed-length feature vector for a fingerprint 
which are invariant under rotation and translation.
This 'test of realness' can be applied to synthetic fingerprints produced by any method.
In this work, tests are conducted on the 12 publicly available databases of FVC2000, FVC2002 and FVC2004 
which are well established benchmarks for evaluating the performance of fingerprint recognition algorithms; 
3~of these 12 databases consist of artificial fingerprints generated by the SFinGe software.
Additionally, we evaluate the discriminative performance 
on a database of synthetic fingerprints generated by the software of Bicz versus real fingerprint images.
We conclude with suggestions for the 
improvement of synthetic fingerprint generation.

\end{abstract}

\section*{Keywords}

Synthetic fingerprint generation, minutiae distribution, transportation problem, earth mover's distance, 
fingerprint reconstruction, identification test, fingerprint individuality.

\clearpage

\begin{figure}
 \begin{center}
   \subfigure[]{\includegraphics[width=0.24\textwidth, height=0.36\textwidth]{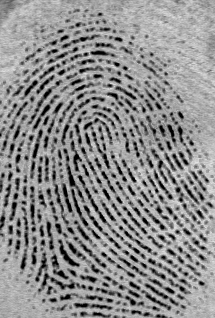}}
   \subfigure[]{\includegraphics[width=0.24\textwidth, height=0.36\textwidth]{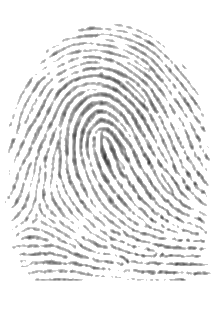}}
   \subfigure[]{\includegraphics[width=0.24\textwidth, height=0.36\textwidth]{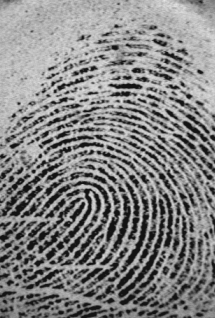}}    
   \subfigure[]{\includegraphics[width=0.24\textwidth, height=0.36\textwidth]{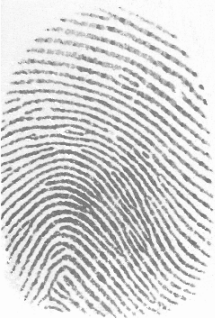}}
 \end{center}
 \caption{Four fingerprint images reproduced from page 292 in \cite{HandbookFingerprintRecognition2009}:
          three of these fingerprint images were acquired from real fingers using different sensors
          and one is a synthetically generated fingerprint. Can you identify the artificial fingerprint? 
          The solution is given 
	  below.
         \label{figSurvey}}
\end{figure}

\textit{At the Fifteenth International Conference on Pattern Recognition held in September 2000,
'about 90 participants, most of them with some background in fingerprint analysis' 
(quoted from page 293 in \cite{HandbookFingerprintRecognition2009})
were shown four fingerpint images (reproduced in Figure \ref{figSurvey}). 
They were told that three images were acquired from real fingers and 
one of these prints was synthetically generated
and they were asked to identify the artificial print.
Only 23\% of the participants correctly chose the synthetic print (image (a) in Figure~\ref{figSurvey}).}

\textit{In this paper, we present a method that is able to perform the task at which the human experts failed:
we introduce a test of realness which can separate real from synthetic fingerprint images
with very high accuracy. We applied the method (a detailed description follows
in Section \ref{Histograms:scn} and \ref{Discrimination:scn}) to these four images 
and the proposed method distinctly identifies the correct image
(computing the difference of minutiae histogram distances explained 
in Section \ref{emdDiff} yields a negative score
for image (a) in Figure \ref{figSurvey} and positive scores for images (b-d)).}

\section{Introduction} 

Today, many commercial, governmental and forensic applications rely on fingerprint recognition 
for verifying or establishing the identity of a person. 
Among these, methods building on minutiae matching play an eminent role. 
Usually, matching routines are not only tested on real fingers, 
but in order to provide for theoretically unlimited sample sizes, 
synthetic fingerprint generation systems such as SFinGe 
by Cappelli \textit{et al.} (\cite{CappelliErolMaioMaltoni2000}) have been developed in the past. 
Independently, methods have been developed to reconstruct fingerprints 
from minutiae templates (cf. \cite{RossShahJain2007}).
Both methodologies are very relevant in many application areas:

\begin{itemize}
 \item Constructing synthetic fingerprint images facilitates the cheap creation 
of very large databases for testing and comparing the performance of algorithms 
in verification and identification scenarios.
 \item Ground truth data is provided for evaluating the performance 
of forensic experts \cite{RodriguezDejonghMeuwly2012} as well as minutiae extraction algorithms. 
 \item Matching performance of low-quality and latent fingerprints is improved~\cite{FengJain2011}.
 \item Fingerprint reconstruction can be a building block 
for solving interoperability problems, e.g. on comparing fingerprints 
acquired from different sensors \cite{HandbookFingerprintRecognition2009}. 
 \item Research in this area raises the awareness for aspects of security, privacy and data protection 
 bearing in mind that an attacker may utilize existing techniques 
for creating a spoof and prepare a presentation attack.
 \item Mixing prints of two or more real fingers has been proposed
for generating virtual identities, obscuring private information 
or creating cancelable templates \cite{OthmanRoss2013}.
\end{itemize}
Of course, synthetic prints should be as real as possible pertaining to all properties and features 
which are relevant for fingerprint recognition, especially with respect to their minutiae distribution. 
Otherwise, a human may be fooled by the look 
of a synthetic print (see the motivational example given before the introduction taken from 
page 293 in \cite{HandbookFingerprintRecognition2009}), 
but their eligibility e.g. for evaluating fingerprint recognition algorithms may be challenged 
and results obtained on artificial databases would be insignificant. 

\paragraph{A unifying concept of the 'correct' minutiae distribution.}
Fingerprint synthesis and fingerprint reconstruction have been treated for a long time as different tasks. 
This can be well conceived on the background that the issue of realistic minutiae distributions 
has only played a subordinate role in theoretical model building and practical research. 
In our contribution we provide for a simple method to assess minutiae distributions of single fingerprints 
as well as of samples of fingerprints. We demonstrate that after training, this allows 
to decide that minutiae patterns of synthetically generated fingerprints are not 'correct'. 
In particular, we believe that including realistic minutiae distributions leads to a unified concept 
in which synthetic fingerprint generation and fingerprint reconstruction
are in fact two sides of the same coin. 
We will return to this point in Section \ref{labelTwoSidesOfTheSameCoin}.

\paragraph{Fingerprints of fingerprints.} 
Minutiae histograms (MHs) introduced below assign any given fingerprint image a fixed length feature vector. 
This feature vector is not only highly potent to discriminate real fingerprints 
from fingerprints synthetically generated by the only publicly available current state of the art system,  
in a preliminary study we additionally demonstrate that this new feature vector 
is also highly discriminatory among real fingerprints as well. 
Given reliably extracted minutiae and sufficent overlap of latent fingerprint images, 
MHs allow for fast and effective matching. 
More precisely, it seems that already the second order MHs reflecting minutiae pairs only,
promise a high potential awaiting to be yet unleashed. 
Note that higher order MHs can be assessed via minutiae triplets, quadruplets etc.

While local minutiae statistics are widely used for matching, our contribution is to consider global minutiae statistics and measure distances by a metric that effectively captures human perception: the earth mover's distance. 

\subsection{Construction and Reconstruction of Fingerprints} \label{secConstructionAndReconstruction}

As made clear above, in our contribution we focus on the role of minutiae distributions. 
This issue is currently gaining  momentum in the scientific community, 
cf. \cite{FengJain2011,KueckenChampod2013,ZhaoJainPaulterTaylor2012}.
In contrast to \cite{ZhaoJainPaulterTaylor2012}, who use parametric mixture models, 
we take a nonparametric statistical approach.

We begin our exposition with biological principles currently 
believed to govern minutiae formation and their distribution. 
To date, however, these principles are still not satisfactorily 
understood \cite{KueckenNewell2004,Kuecken2007,KueckenChampod2013}. 
Subsequently, we put the SFinGe method into context with these biological hypotheses 
and describe an alternate method based on real fingerprint images for generating latent fingerprints, 
contrasting in this and other aspects. 
Thereafter, we allude to the similarities between fingerprint generation and reconstruction 
which will resurface when we propose improvements in synthetic fingerprint generation. 
We conclude this section by laying out the plan of the paper.

\paragraph{Fingerprint formation guided by Merkel cells.}

K\"ucken and Champod propose a model for fingerprint formation \cite{KueckenChampod2013} 
that has two major influence factors: 
growth forces which create mechanical compressive stress \cite{KueckenNewell2004,KueckenNewell2005}
on the one hand, and Merkel cells rearranging from a random initial configuration 
into lines minimizing the compressive stress and inducing primary ridges on the other hand.
Merkel cells interact with each other in reaction-diffusion systems of short range attraction 
and long range repulsion~\cite{GiererMeinhardt1972}. Based on empirical evidence 
from embryonic volar tissue evolution (e.g. \cite{Babler1991}), 
K\"ucken and Champod let solutions of suitable partial differential equations propagate 
from three centers: one along the flexion crease, one at the volar pad (the core area) and one from the nail furrow. 
Based on specific parameter choices, this process eventually forms a ridge line pattern featuring minutiae. 
K\"ucken and Champod analyzed the images resulting from simulation runs of their growth model 
and discovered qualitative differences in comparison to real fingerprints. 
They conclude that with respect to the natural variability of arrangements 
of minutiae 'only an empirical acquisition of genuine fingerprints will provide 
an adequate source of data' \cite{KueckenChampod2013}.

\paragraph{SFinGe.}

In a nutshell, growing fingerprint patches containing no minutiae are generated starting from a number of randomly located points by the iterative application of Gabor filters according to 
a previously generated global orientation field \cite{VizcayaGerhardt1996}
and a non-constant ridge frequency pattern.
Whenever patches meet, minutiae are produced whenever necessary for the consistency of the global ridge pattern. 
A detailed description of the SFinGe method by Cappelli \textit{et al.} can be found in \cite{CappelliErolMaioMaltoni2000}, 
Chapter 6 in \cite{HandbookFingerprintRecognition2009} and Chapter 18 in \cite{AutomaticFingerprintRecognitionSystems2004}.
 
In fact, the SFinGe model silently assumes biological hypotheses of fingerprint pattern formation that are slightly different from the ones described above. First, fingerprint patterns no longer propagate from a well defined system of three original sources but rather from 
a multitude of sources at random locations. Secondly, a main governing principle for minutiae creation lies in the compatibility of ridge patterns whenever growing patches touch. These hypotheses in itself are very interesting and can be viewed as intuitively natural; and their validity can be assessed with our methodology. 
Our test of realness (Section \ref{secTestRealness}) yields, however, that these hypotheses explain 
the process of minutiae formation not satisfactorily (Section \ref{secResults}).

Notably, this shortfall of SFinGe cannot be explained by imprint distortions (which are already included in the SFinGe model) and the lack of a true 3D model, as both would account mostly for neighboring bin perturbations or none at all. 
Using the earth mover's distance \cite{GottschlichSchuhmacher2014}, precisely distinguishes 
between these minor perturbations and the major deviations we found (for a detailed discussion cf. \cite{RubnerTomasiGuibas2000}).

\paragraph{Further methods.}

\begin{figure}
 \begin{center}
   \subfigure[]{\includegraphics[width=0.40\textwidth, height=0.40\textwidth]{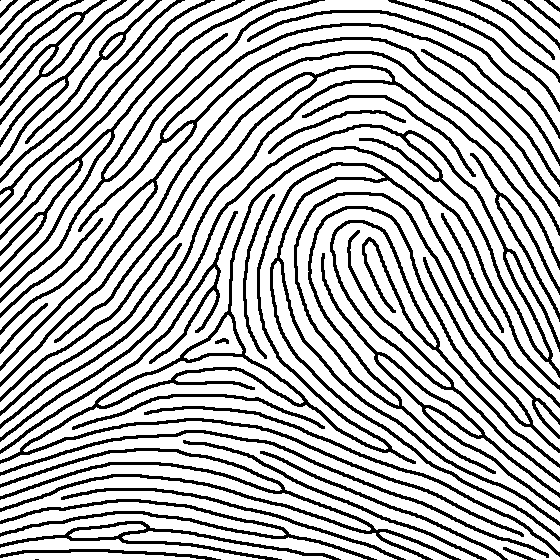}}
   \subfigure[]{\includegraphics[width=0.40\textwidth, height=0.40\textwidth]{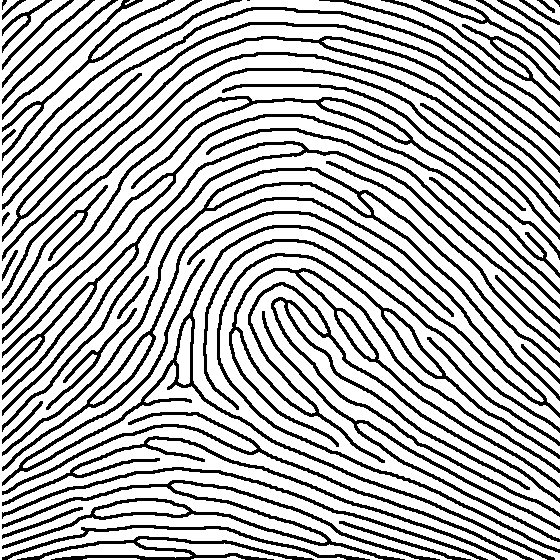}} \\
   \subfigure[]{\includegraphics[width=0.40\textwidth, height=0.40\textwidth]{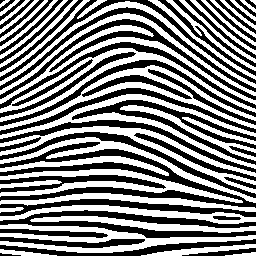}}
   \subfigure[]{\includegraphics[width=0.40\textwidth, height=0.40\textwidth]{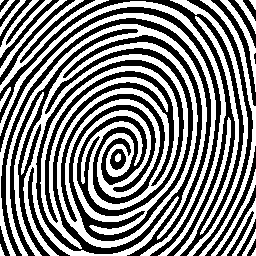}} \\
   \subfigure[]{\includegraphics[width=0.40\textwidth, height=0.40\textwidth]{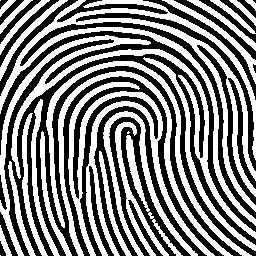}}    
   \subfigure[]{\includegraphics[width=0.40\textwidth, height=0.40\textwidth]{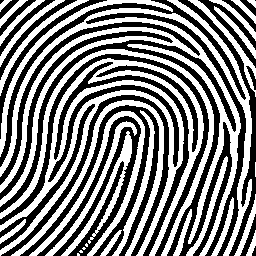}} 
 \end{center}
 \caption{(a) and (b) show examples of synthetic fingerprints generated 
          by the method of Araque~et~al.~\cite{AraqueEtAl2002};
          (c) to (f) display synthetic fingerprint images created 
          by the software of Bicz~\cite{Bicz2003}.
         \label{figSynthetic}}
\end{figure}

Further methods for generating synthetic fingerprint images were proposed 
in 2002 by Araque et al. \cite{AraqueEtAl2002}
and in 2003 by Bicz~\cite{Bicz2003}. 
The methods of Araque et al. and SFinGe are similar in spirit
and they both rely on the global orientation field model of Vizcaya and Gerhardt \cite{VizcayaGerhardt1996}
for creating an orientation field. 
Examples of synthetic fingerprints generated by this method are shown in Figure \ref{figSynthetic} (a) and (b).

Viewing fingerprints as holograms (cf. \cite{LarkinFletcher2007}), a given global orientation field yields the continuous phase onto which so called spiral phases can be added giving minutiae at specific predefined locations while other minutiae occur due to continuity constraints. This allows for synthetic fingerprint generation and four example images generated by the software of Bicz~\cite{Bicz2003} 
are displayed in Figure \ref{figSynthetic} (c) to (f).
Results reported in Section \ref{secResults} show that 
the method proposed in this paper can also discriminate 
between synthetic fingerprints according to Bicz and real fingerprint images.

\paragraph{Creating forensic fingermarks from real prints.}

The approach by Rodriguez \textit{et al.} \cite{RodriguezDejonghMeuwly2012}, 
since the minutiae are based on real fingers, 
does not suffer from unrealistic minutiae configurations. 
It is however subject to increased time, money and data protection constraints.
Forensic fingermarks (latent fingerprint images) 
are simulated in a semi-automatic way: 
fingers of volunteers are recorded using a livescan device.
During a period of about 30 seconds, each person performs a series of predefined movements 
which results in fingerprint images with various distortions. 
Images are captured at a rate of four frames per second.
Minutiae are automatically extracted followed by manual inspection and correction
of falsely extracted or missed minutiae. 
Starting with this ground truth data set, 
latents are simulated using a region containing a cluster of 5-12 minutiae
from the real fingerprint images.

One use case for these simulated latents is to test 
the minutiae marking performance of human experts
which can be evaluated against the avaivable ground truth information.
Another application area is benchmarking the identification performance of AFIS software.

\paragraph{Reconstruction.}

Various researches have shown that fingerprints 
can be reconstructed from minutiae templates \cite{CappelliLuminiMaioMaltoni2007,RossShahJain2007,FengJain2011,LiKot2012}.
There are two major approaches for the automatic reconstruction of fingerprint images:
first, the iterative application of Gabor Filters as in SFinGe, and second, 
the usage of amplitude- and frequency-modulated (AM-FM) functions \cite{Larkin2001,LarkinBoneOldfield2001,LarkinFletcher2007}.
In both cases, the first step is the estimation of an orientation field 
which fits the minutiae pattern.
Fingerprint reconstruction from minutiae templates 
and the generation of synthetic images follow similar principles.
However, the goal in the reconstruction scenario is 
to produce the same number of minutiae at the same locations 
and with same direction and type as in the template. 

\paragraph{Synthetics vs. spoofs.} The MH based method proposed in this paper does not perform 
liveness detection \cite{GottschlichMarascoYangCukic2014,ChoiKangChoHinKim2009,JinBaeMaengLee2010,SousedikBusch2014,MarascoRoss2014}.
In fact, we would like to stress that if an attacker can produce a high quality spoof containing the same minutiae 
as the imitated alive finger and puts this spoof finger on fingerprint sensor,
the proposed method will classify the acquired fingerprint image as originating from a real finger
with respect to the minutiae distribution. 
Only attacks with spoofs based on synthetic images and synthetic minutiae distributions are addressed in this work.

\subsection{Plan of the Paper}
In the next section, we introduce second order extended minutiae histograms (MHs)
which for this paper we just call minutiae histograms:
statistics of pairs of two minutiae are used in combination 
with minutiae type information and interridge distances to obtain a fingerprint of fingerprints.
In Section \ref{Discrimination:scn}, extended MHs are applied for classifying 
a fingerprint into one of the two categories, real or synthetic. 
Tests on the 12 publicly available databases of FVC2000, FVC2002 and FVC2004 
show the discriminative power of this approach. 
In Section \ref{secImprovement}, detailed suggestions for the generation 
of more realistic synthetic fingerprints are given.
In Section \ref{secIdentification}, the suitability of minutiae histograms for identification purposes is investigated
and in Section \ref{secIndividuality}, MHs are proposed for the quantification of fingerprint evidence.
Section \ref{Discussion:scn} conludes with a discussion and states topics of further research.

\begin{figure}
 \begin{center}
   \subfigure[]{\includegraphics[width=0.40\textwidth, height=0.45\textwidth]{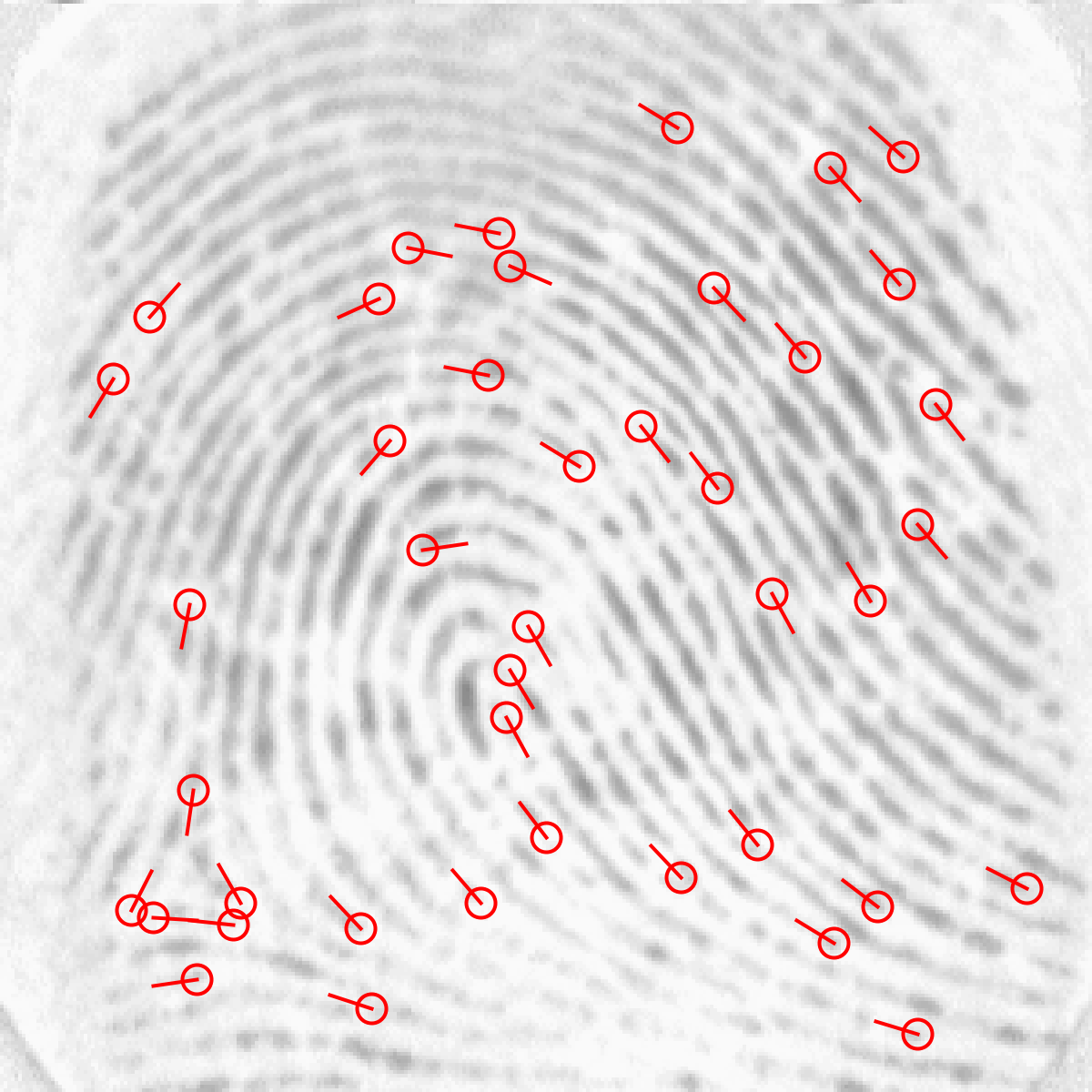}}
   \subfigure[]{\includegraphics[width=0.40\textwidth, height=0.45\textwidth]{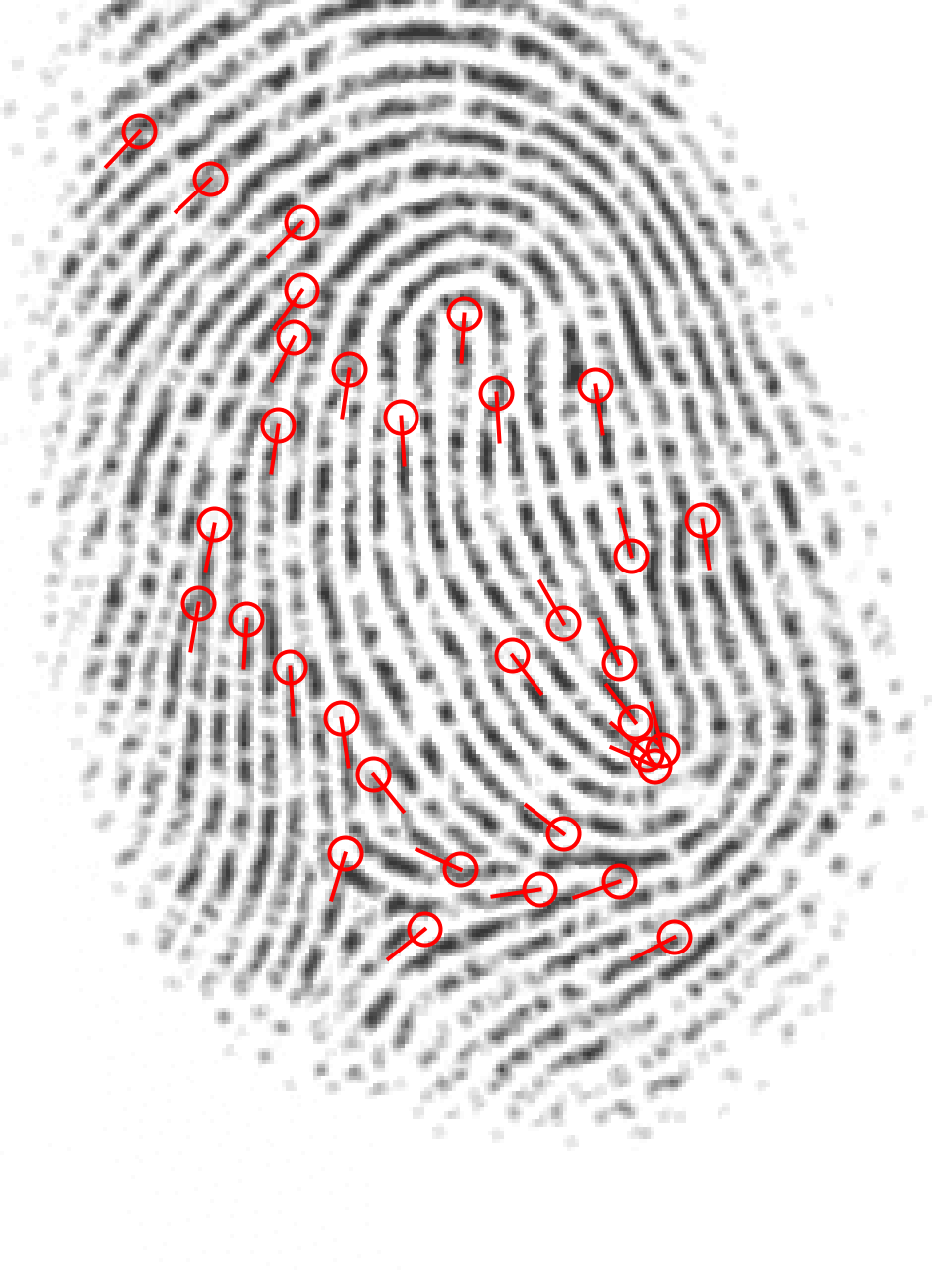}} \\
   \subfigure[]{\includegraphics[width=0.20\textwidth, height=0.20\textwidth]{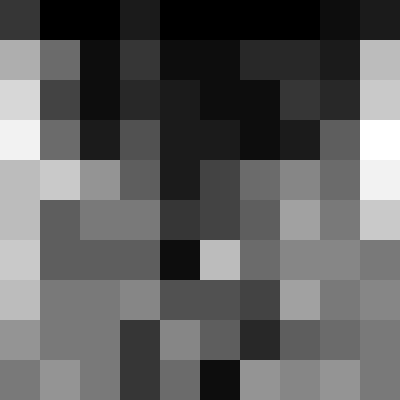}}
   \subfigure[]{\includegraphics[width=0.20\textwidth, height=0.20\textwidth]{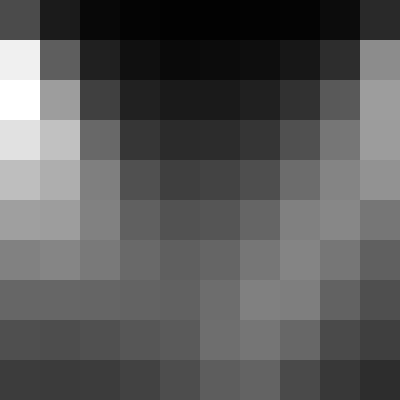}}
   \subfigure[]{\includegraphics[width=0.20\textwidth, height=0.20\textwidth]{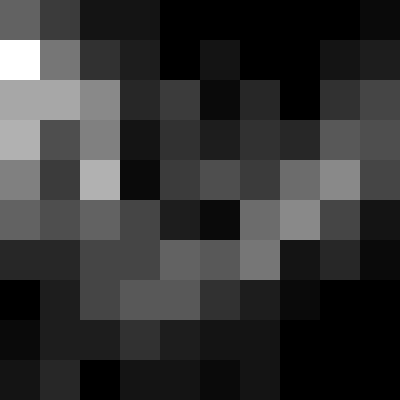}}    
   \subfigure[]{\includegraphics[width=0.20\textwidth, height=0.20\textwidth]{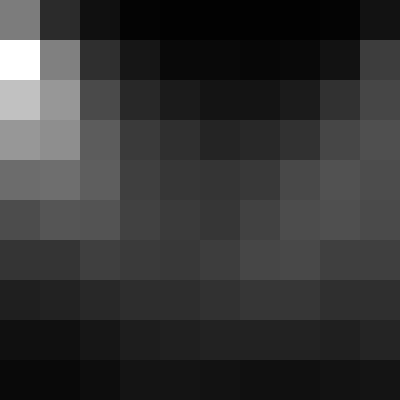}}
 \end{center}
 \caption{For a detailed description of the underlying sets I and III cf. Section \ref{Results:scn}. 
         (a) shows a minutiae template extracted from a real finger of FVC2000 DB1 Set III
         and (b) from a synthetic print of FVC2000 DB4.
         (c) visualizes the derived 2D-MH for the template in (a),
         and correspondingly (e) displays the histogram of (b).
         The average histogram of FVC2000 DB1 Set I is visualized in (d) 
         and for the synthetic fingers of FVC2000 DB4 in (f).
         The EMD (see Section \ref{secMH}) from (c) to (d) is 0.66 and (c) to (f) 1.79.
         Therefore, the template is correctly classified as belonging to a real finger.
         The EMD from (e) to (d) is 1.69 und from (e) to (f) is 0.61,
         and consequently, (e) is correctly recognized as stemming from an artificial finger.
         In (c-f), distance bins are displayed from top to bottom, 
         directional difference bins from left to right.
         A high brightness value corresponds to a high number of occurrences in a bin.
         \label{figTemplate}}
\end{figure}

\begin{figure} 
 \begin{tabular}{|c|c|c|} \hline
  & real & synthetic \\ \hline 
  & & \\
  nearby angles & \includegraphics[width=0.3\textwidth]{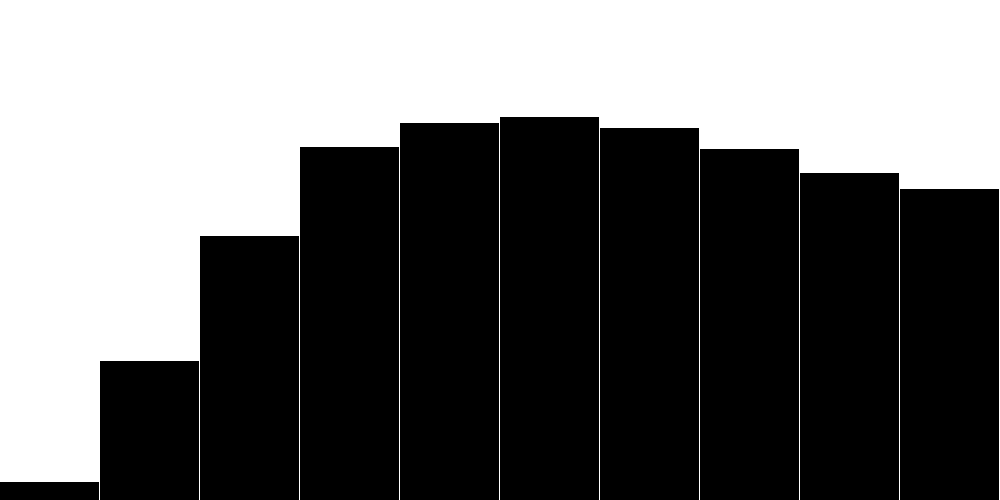} 
                & \includegraphics[width=0.3\textwidth]{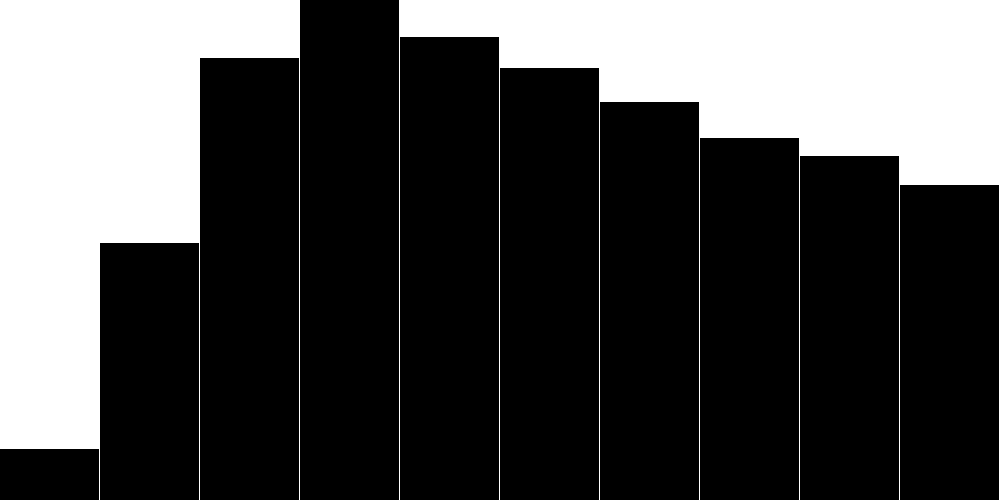}\\ \hline 
  & & \\
  nearly opposite angles & \includegraphics[width=0.3\textwidth]{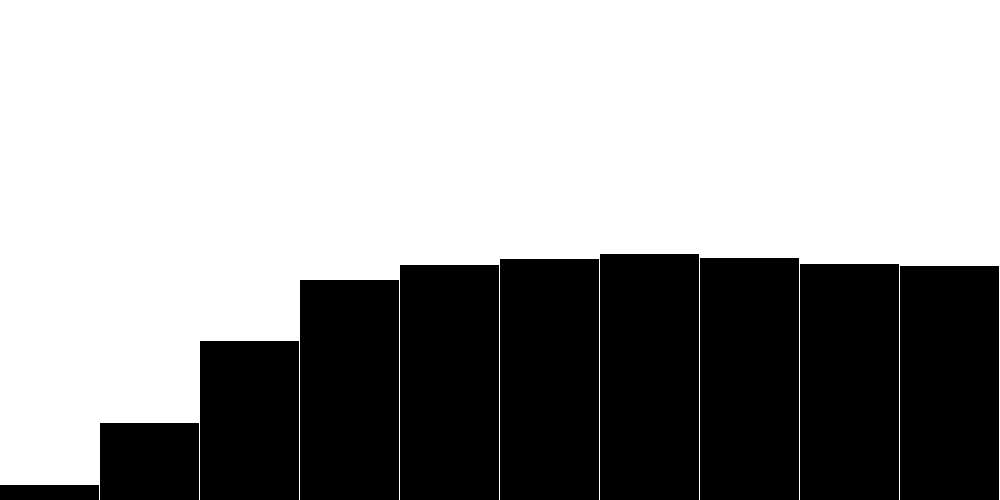} 
                         & \includegraphics[width=0.3\textwidth]{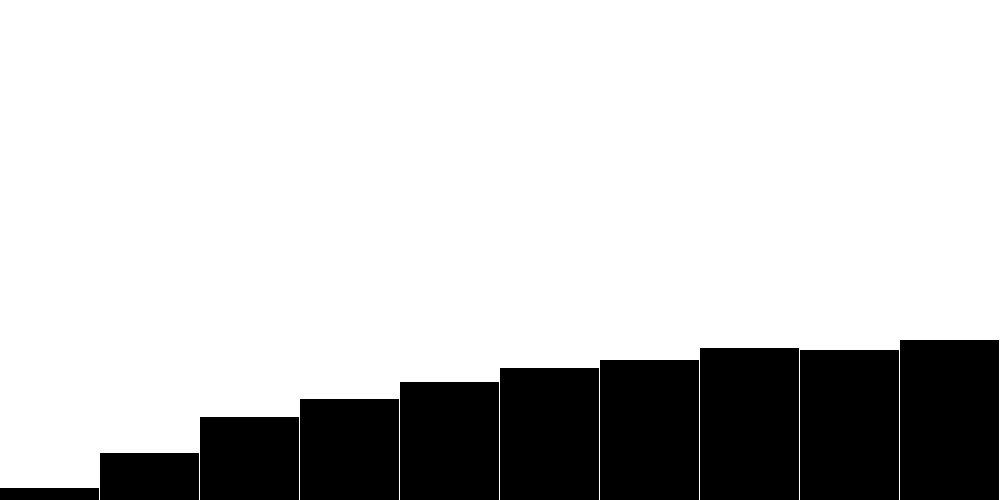}\\ \hline
 \end{tabular}
 \caption{\it Details of mean MHs displayed for specific angles and mutual minutiae distances 
between 0 and 80 pixels (each bin is 8 pixels wide) over the databases of FVC2000, FVC2002 and FVC2004. 
Left column: means of real fingers (DB1, DB2 and DB3 combined). 
Right colunn: mean over synthetic fingerprints simulated with SFinGe (DB4). 
Top row: counts of minutiae of mutual absolute angle distance $\in [0,18^o]$. 
Bottom row: counts of minutiae of mutual absolute angle distance $\in [172^o,180^o]$.
The height of the bar corresponds to the mean percentage of minutiae pairs sorted into that bin.
\label{figHistogram1DAll}}
\end{figure}

\section{Extended Minutiae Histograms}\label{Histograms:scn}

Minutiae histograms (MHs) feature relations between minutiae. Here we consider only second order MHs which feature relations between minutiae pairs. Higher order MHs feature relations between triples, quadruplets, etc., respectively. As detailed below, in the following we consider only distance and directional relations giving 2D-MHs. Including the entire spatial relations and minutiae type would give 4D-MHs which are briefly discussed further down. Moreover, MHs can be extended by adding other relevant features giving \emph{extended minutiae histograms} comprising
\begin{itemize}
 \item[(i)] 2D-MHs or 4D-MHs featuring a histograms of minutiae pairs,
 \item[(ii)] interridge distances and
 \item[(iii)] the percentage of bifurcations and endings in templates
\end{itemize}
cf. Table \ref{tabFeat}.

\begin{table}[ht!]
\begin{center}
\begin{tabular}{|c|l|} \hline
$d$ & distance in pixels between two minutiae. \\ \hline
$\alpha$ & directional difference between two minutiae directions.\\ \hline
$m_{IRD}$ & global mean interridge distance of a fingerprint.\\ \hline
$v_{IRD}$ & global variance of interridge distances.\\ \hline
$p_{bif}$ & percentage of bifurcations in a template.\\ \hline
\end{tabular}
\caption{Extended 2D-MH feature overview. 
         \label{tabFeat}}
\end{center}
\end{table}

\subsection{2D-Minutiae Histograms}
\label{secMH}

Here, we consider 2D-minutiae histograms, 
because they carry sufficient power to discriminate between real and synthetic prints.
We construct a two-dimensional frequency histogram 
by computing the distance $d$ between minutiae locations in pixels and 
directional difference $\alpha$ in degrees of the two minutiae directions 
for all combinations of two minutiae in a template.
Both features are binned using identically sized, equidistant intervals.

Figure \ref{figTemplate} (c) to (f) shows four histograms with $10 \times 10$ bins.
Here, distances $d_i \leq d_{max} = 200$ pixels are divided into intervals of 20 pixels ranges
(distances increase from top to bottom) 
and directional differences $\alpha_i$ into 10 bins of $180\degree$ total range. 
Each directional difference bin consists of two intervals of $18\degree$ range and
differences $> 180\degree$ are mirrored into the corresponding bin,
e.g., $\alpha_1 = 10\degree$ and $\alpha_2 = 350\degree$ are portioned into the same bin,
$\alpha_3 = 170\degree$ and $\alpha_4 = 190\degree$ are grouped into one bin 
and $\alpha_5 = 90\degree$ and $\alpha_6 = 270\degree$ are also indexed into the same bin.
In Figure \ref{figTemplate} (c) to (f), 
the bins of first column on the left are centered at $0\degree$.
A detail of the first and last column is displayed in Figure \ref{figHistogram1DAll}.
Highlighting the differring distributions, we have chosen $d_{max} = 80$ pixels.

In order to obtain a sensor-independent and age-independent descriptor,
the minutiae templates are rescaled to prints at 500 DPI.
The templates of FVC2002 DB2 (569 DPI) and FVC2004 DB3 (512 DPI) are demagnified accordingly
and the rescaling is taken into account during the estimation of the interridge distances listed 
in Table \ref{tabIR}. The rescaling facilitates a fair comparison with synthetic prints
which are designed to produce fingerprint images of approximately 500 DPI.

The description of FVC2000 DB3 \cite{FVC2000} states 
that one third of the 19 volunteers were under 18 years of age.
These prints could be easily enlarged to the predicted adult print size at 500 DPI
using the scaling factor attained in \cite{GottschlichHotzLorenzBernhardtHantschelMunk2011},
if additional information were available (the age and whether the fingerprints was acquired from 
a male or female person). Since children and adolescents have smaller fingers, 
the average ridge frequency and the distances between two minutiae are smaller and 
applying the same bin limits as for adults 
to the unscaled templates of still growing persons results in different statistics.
As a consequence, the classification could be even better, if this additional information was available.

\begin{table}[ht!]
\begin{center}
\begin{tabular}{|c|l|} \hline
$d_{max}$ & Maximal distance in pixels between two minutiae. \\ \hline
$b_{dist}$ & {\#} of bins for distances in pixels.\\ \hline
$b_{dir}$ & {\#} of bins covering differences between two minutiae dirctions.\\ \hline
${r}$ & Costs for moving mass along neighboring directional bins.\\ \hline
${s}$ & Costs for moving mass along neighboring distance bins.\\ \hline
${e}$ & Exponentiation factor for costs.\\ \hline
\end{tabular}
\caption{2D-MH parameter overview. 
         \label{tabParam}}
\end{center}
\end{table}

The distance between two 2D-MHs is computed by the earth mover's distance (EMD) \cite{GottschlichSchuhmacher2014}
which measures the distance between two distributions 
by computing the minimal cost for transforming one distribution into the other.
This metric is especially useful for comparing histograms~\cite{LingOkada2007}
and has many applications, e.g. in content-based image retrieval \cite{RubnerTomasiGuibas2000}.
The concept was first described by G. Monge in 1781 and is also known as 
Mallows distance, Wasserstein distance or Kantorovich-Rubinstein distance.

The cost matrix of the transport problem consists of the composite costs for 
moving mass along the distance bins and for moving mass 
along the directional difference bins. 
For example, moving mass $m$ from distance bin $x$ und directional bin $u$ 
to distance bin $y$ and directional bin $v$ results in the following costs $c$:
\begin{equation*}
c = m \cdot (  ({s} \cdot |x-y|)^{e} + ({r} \cdot |u-v|)^{e} )   
\end{equation*}
The auction algorithm of Bertsekas \cite{Bertsekas1991}
or the shortlist method \cite{GottschlichSchuhmacher2014}
can be applied to solve the transportation problem and to compute the EMD.
The favorable computational runtime of the shortlist method 
in comparison to other methods gains relevance especially
for transportation problems of larger dimensionality,
or if a very large number of EMDs is to be computed~\cite{GottschlichSchuhmacher2014}.

For the classification of fingerprints into real or synthetic, 
an average 2D-MH is computed for each of the two classes 
(using a set of minutiae templates described in Section~\ref{Discrimination:scn}).
The average 2D-MHs act as representatives for its respective class.
All 2D-MH are normalized such that the sum of the masses in all bins amounts to~1.

\label{emdDiff}

For a minutiae template which shall be classified into real or synthetic,
first, the 2D-MH of the template is computed and it is normalized.
Next, the EMD between the 2D-MH of the unknown class and 
the average 2D-MH of the real fingers is computed as well as 
and the EMD to the average 2D-MH of the synthetic prints.
The minutiae template can be classified by simply choosing the class with the smaller EMD 
(see upper half in Table \ref{tabPerformanceOverview}) 
or both EMDs can used in combination with additional features 
(see lower half in Table \ref{tabPerformanceOverview}).
The score referred to in Figure \ref{figSurvey} is simply the difference of the EMDs.
For identification, we use unnormalized intersection distances 
of unnormalized MHs instead, cf. Section \ref{secIdentification}.

\subsection{4D-MHs}

Minutiae properties not considered so far in the 2D-MH
are the angle of the relative position of the second minutia 
with respect to the first and the local combination of minutiae types.
Minutiae types are currently accounted for 
only as a global value (percentage of bifurcations in a template).

In order to classify a print as real or synthetic, 
this additional information is not necessary, 
but for other applications, it can be useful and should be considered.
The 2D-MHs can easily be augmented by these additional two dimensions giving 4D-MHs.

In Section \ref{secIdentification}, the usage of 4D-MHs 
in an identification scenario is studied 
and in Section \ref{secIndividuality}, we propose to apply 
the empirical distribution of extended 4D-MHs for quantifying 
the weight of fingerprint evindence in court.

\subsection{Minutiae Type}

\begin{table}[ht!]
\begin{center}
\begin{tabular}{|c|c|c|c|c|} \hline
        & DB 1 & DB 2 & DB 3 & DB 4 \\ \hline
FVC2000 & 38.8 & 43.6 & 40.2 & 30.0 \\ \hline
FVC2002 & 38.1 & 37.8 & 36.3 & 29.2 \\ \hline
FVC2004 & 46.5 & 37.3 & 55.8 & 32.1 \\ \hline 
\end{tabular}
\caption{Average percentages of bifurcations in minutiae templates 
         for real (DB 1-3) and synthetic (DB 4) fingerprints.
         \label{tabPercentBifurcations}}
\end{center}
\end{table}

Table \ref{tabPercentBifurcations} lists the average percentages of bifurcations 
in minutiae templates for each of the FVC databases. 
Some fluctations among the databases containing real fingerprints can be observed 
which can be attributed to various causes including image quality 
and its influence on the automatic minutiae extraction, 
the usage of different sensors with different properties 
including the size of the captured finger surface, 
the variability and distribution of minutiae in the fingers of the volunteers 
whose fingerprints were acquired for building the databases. 
However, we notice that the artificial fingerprints generated 
by SFinGe tend to have systematically lower percentages 
of bifurcations which is presumably caused by the image fabrication process.
Therefore, this percentage is included as a feature 
for the proposed classification of a fingerprint as real or synthetic.

\subsection{Interridge Distances}

\begin{figure}
 \begin{center}
   \includegraphics[width=0.9\textwidth]{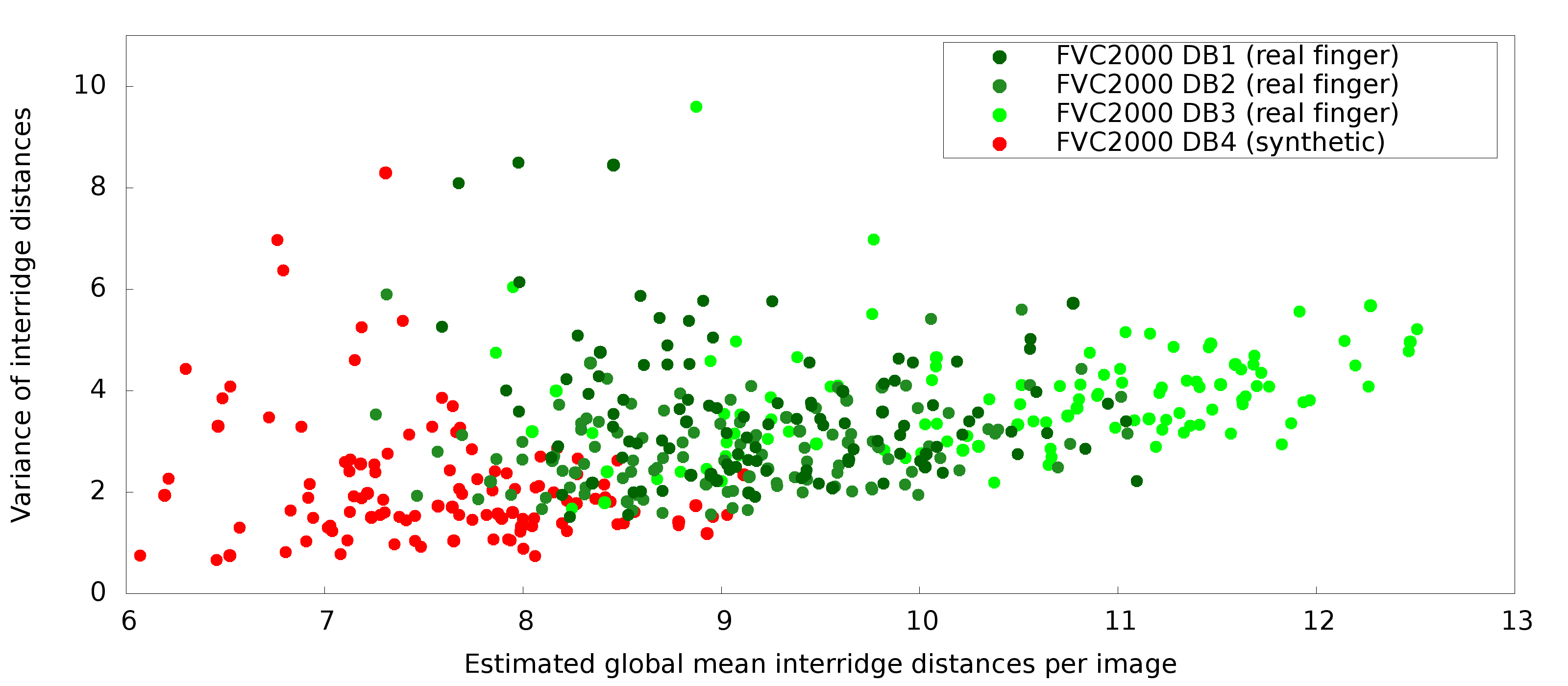} \\
   \includegraphics[width=0.9\textwidth]{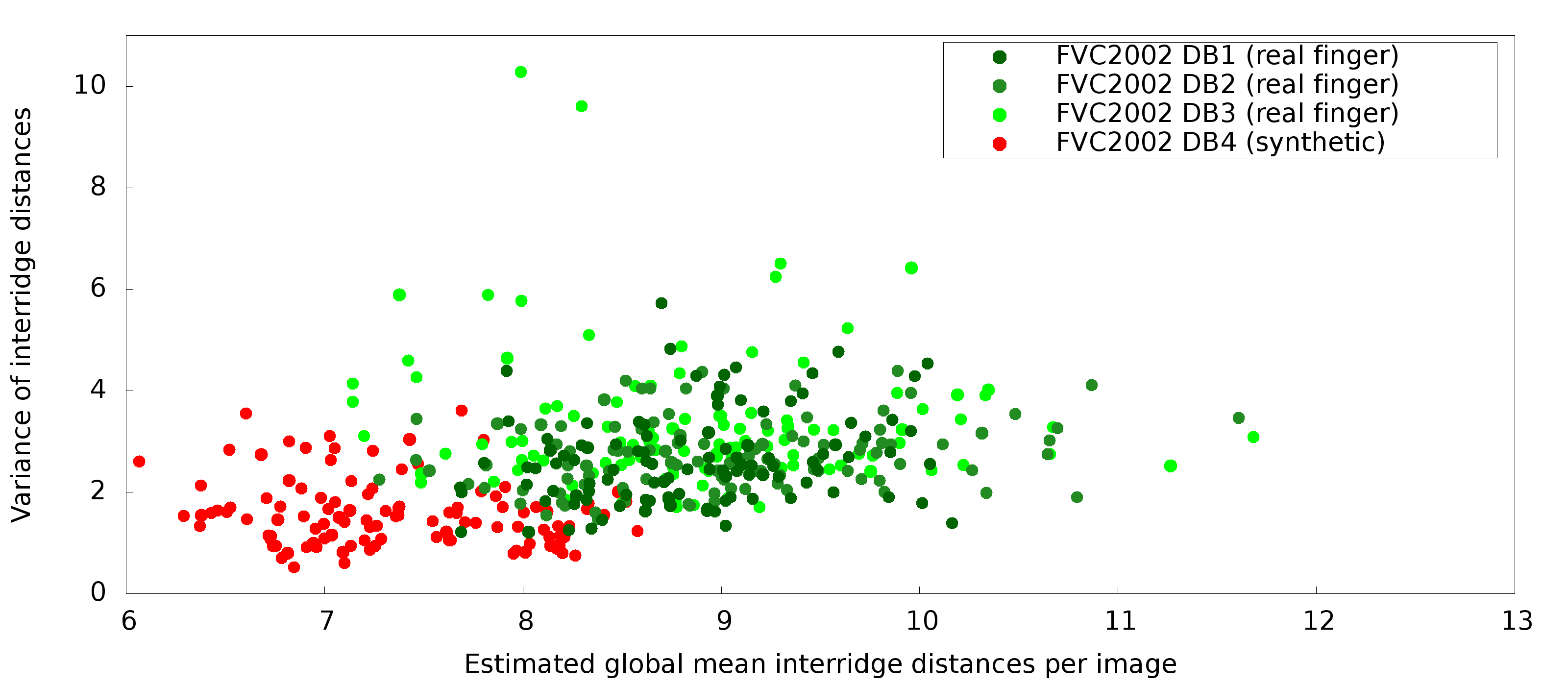} \\
   \includegraphics[width=0.9\textwidth]{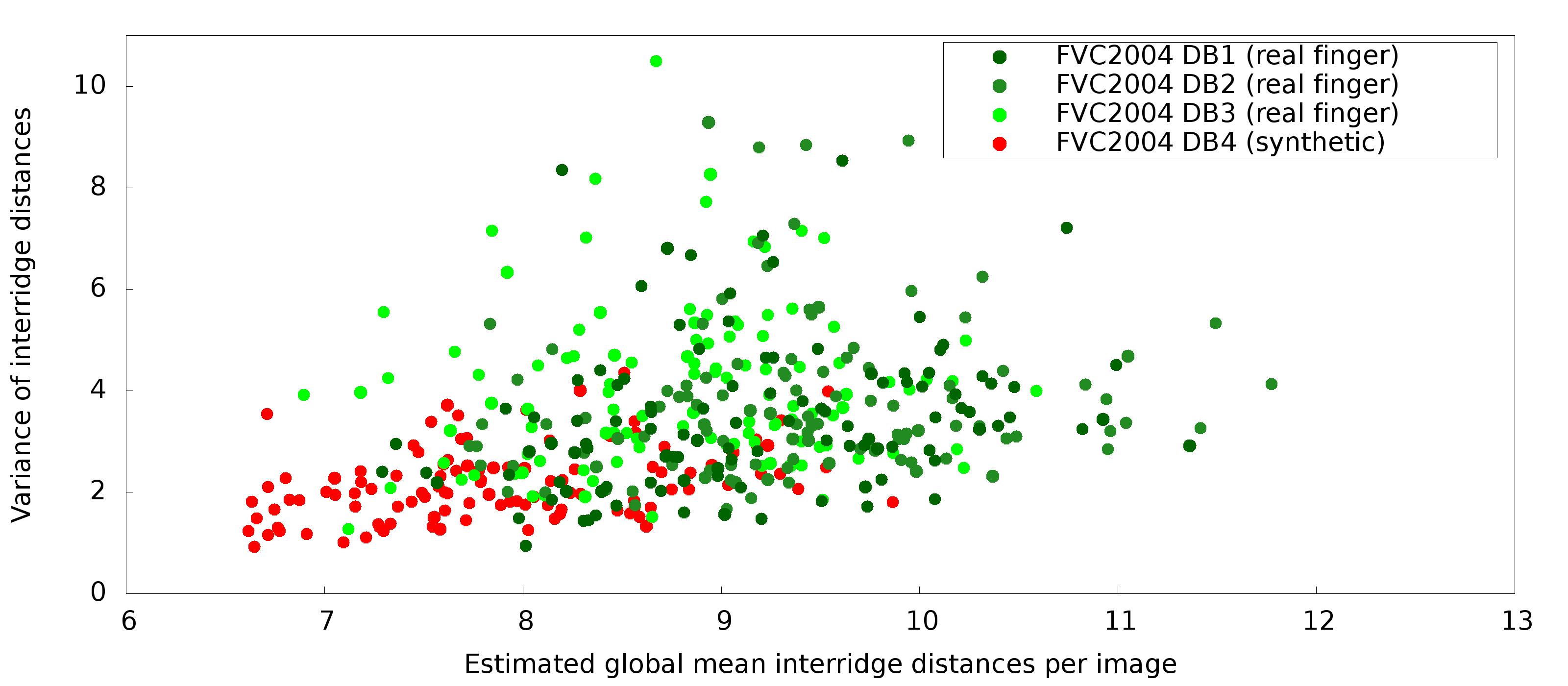}
 \end{center}
 \caption{Global mean and variance of individual interridge distances per image for all images 
          of FVC2000, FVC2002 and FVC2004 estimated by curved regions \cite{Gottschlich2012}.
          Systematic differences between images acquired from real fingers (green) 
          and synthetic images (red) are clearly visible.
         \label{figIRD}}
\end{figure}

The interidge distance image $\Xi$ assigns each pixel a local estimate 
of the distance between two neighboring ridges (or two valleys). 
For prints acquired from adults at a resolution of 500 DPI, 
the value of $\Xi(x,y)$ is in the range from 3 to 25 pixels (forensic researchers 
have shown that women tend to have on average slightly smaller 
interridge distances than men (see \cite{GutierrezrodomeroAlonsoRomeroGalera2008,NayakRastogiKanchanYoganarasimhaKumarMenezes2010})).

The interridge distances listed in Table \ref{tabIR} are estimated 
by curved regions as described in \cite{Gottschlich2012}.
The orientation field estimate for constructing the curved regions is obtained 
by using a combination of the line sensor method \cite{GottschlichMihailescuMunk2009} 
and a gradients based method as described in \cite{GottschlichSchoenlieb2012}.

We observe (see Table \ref{tabIR}) that the synthetic fingerprints (DB4)
are marked by lower estimated mean interridge distances 
and a lower variance of the interridge distances in comparison 
to images of real fingers (DB1-DB3).
Figure \ref{figIRD} shows the distribution of the global mean interidge distances and the global variance of interridge distances for all images of FVC2000, FVC2002 and FVC2004.

\begin{table}[ht!]
\begin{center}
\begin{tabular}{|c|c|c|c|c|c|c|c|c|} \hline
&\multicolumn{4}{c|}{Mean interridge distance}&\multicolumn{4}{c|}{Variance} \\ \hline
&DB1&DB2&DB3&DB4&DB1&DB2&DB3&DB4 \\ \hline
FVC2000& 9.24 &  9.07 & 10.53 & 7.60 & 3.53 & 2.87 & 3.85 & 2.15 \\ \hline
FVC2002& 8.85 &  9.00 &  8.88 & 7.34 & 2.63 & 2.73 & 3.35 & 1.58 \\ \hline
FVC2004& 9.15 &  9.40 &  8.80 & 7.90 & 3.46 & 3.86 & 4.06 & 2.17 \\ \hline
\end{tabular}
\caption{Mean and variance of interridge distances for real (DB1 - DB3) and synthetic (DB4) fingerprints.
         \label{tabIR}}
\end{center}
\end{table}

\section{Separating the Real from the Synthetic}\label{Discrimination:scn}

\subsection{The Test of Realness: Training and Test Protocol}

\label{secTestRealness}

Tests are conducted on the publicly avaivable fingerprint competitions 
of FVC2000, FVC2002 and FVC2004 \cite{FVC2000, FVC2002, FVC2004}.
Each competition consists of four databases (DB1-DB4): 
for the first three databases, fingerprints were acquired from volunteers using different sensors.
The fourth database contains synthetic fingerprints created by the SFinGe software. 

All databases with real and synthetic prints contain images from 110 fingers with 8 impressions per finger.
We divided each set of 110 fingers into three independent, non-overlapping sets (see Table \ref{tabSets}).

\begin{table}[ht!]
\begin{center}
\begin{tabular}{|c|l|l|} \hline
Set I & Finger 1 to 40   & Computing an average template \\ \hline
Set II & Finger 41 to 70  & Parameter training\\ \hline
Set III & Finger 71 to 110 & Testing the classification performance \\ \hline
\end{tabular}
\caption{Set overview for each competion and database.
         \label{tabSets}}
\end{center}
\end{table}

On set I, the average template is computed which acts as a representative for its class 
(real for DB 1-3 and synthetic for DB 4).
A few combinations of the parameters listed in Table \ref{tabParam} 
and weights for linear feature fusion are trained on set II
and the configuration which leads to the best classification is chosen for the test on set III.

The features are fused into a combined score $s$ which is computed as 
$s = w_0 + w_1 \cdot a + w_2 \cdot b + w_3 \cdot c + w_4 \cdot d$,
where 
$a$ is the histogram EMD difference, 
$b$ the normalized global mean of interridge distances,
$c$ the normalized global variance of interridge distances and
$d$ the normalized percentage of bifurcations in a template.  
The weights $w_i$ are trained on set II.

\begin{table}[ht!]
\begin{center}
\begin{tabular}{|c|c|c|c|c|c|c|c|c|c|c|c|c|} \hline
\multirow{3}{*}{} & \multicolumn{4}{c|}{{FVC2000}} & \multicolumn{4}{c|}{{FVC2002}} & \multicolumn{4}{c|}{{FVC2004}} \\ \cline{2-13}
 &  DB1 & DB2 & DB3 & DB4 & DB1 & DB2 & DB3 & DB4 & DB1 & DB2 & DB3 & DB4\\ \cline{2-13}
 &  A & B & C & D & E & F & G & H & I & J & K & L \\ \hline
A& 0    & 0.02 & 0.06 & 1.11 
 & 0.03 & 0.03 & 0.10 & 0.68 
 & 0.03 & 0.05 & 0.04 & 0.44 \\ \hline
B&      & 0    & 0.05 & 1.11   
 & 0.03 & 0.04 & 0.11 & 0.68 
 & 0.02 & 0.05 & 0.03 & 0.43 \\ \hline
C&      &      & 0    & 1.10 
 & 0.06 & 0.08 & 0.10 & 0.68 
 & 0.05 & 0.03 & 0.03 & 0.44 \\ \hline 
D&      &      &      & 0 
 & 1.11 & 1.11 & 1.12 & 0.58
 & 1.11 & 1.11 & 1.11 & 0.81 \\ \hline
E&      &      &      &   
 & 0    & 0.03 & 0.11 & 0.68
 & 0.02 & 0.06 & 0.04 & 0.44 \\ \hline
F&      &      &      &  
 &      & 0    & 0.11 & 0.68
 & 0.04 & 0.07 & 0.06 & 0.44 \\ \hline
G&      &      &      &  
 &      &      & 0    & 0.71  
 & 0.11 & 0.08 & 0.10 & 0.48 \\ \hline
H&      &      &      & 
 &      &      &      & 0 
 & 0.68 & 0.69 & 0.69 & 0.29 \\ \hline
I&      &      &      &  
 &      &      &      &   
 & 0    & 0.05 & 0.04 & 0.43 \\ \hline
J&      &      &      &  
 &      &      &      &  
 &      & 0    & 0.03 & 0.45 \\ \hline
K&      &      &      &  
 &      &      &      & 
 &      &      & 0    & 0.45 \\ \hline
L&      &      &      & 
 &      &      &      &  
 &      &      &      & 0\\ \hline
\end{tabular}
\caption{EMDs between the average 2D-MHs per database for all combinations 
         of databases (cf. Figure \ref{MinutiaeAngleDist-hist-WassersteinMDS:fig}). 
         \label{tabAvgEMD}}
\end{center}
\end{table}

\begin{figure}
\centering
  \subfigure[]{\includegraphics[width=0.45\textwidth]{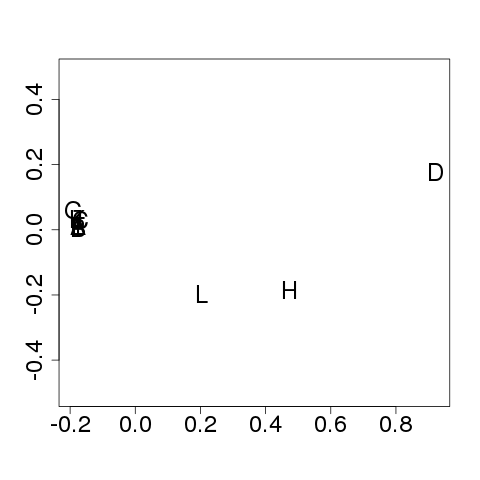}}
  \subfigure[]{\includegraphics[width=0.45\textwidth]{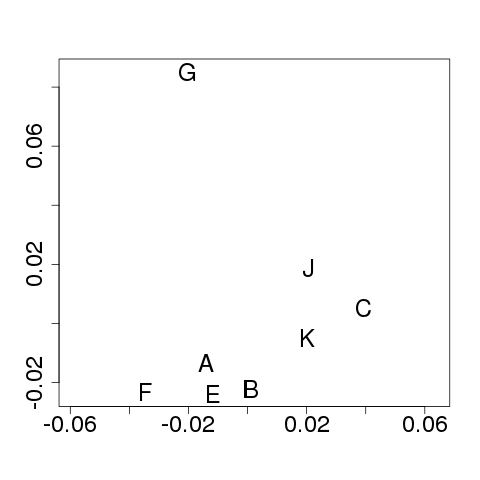}}
 \caption{Two-dimensional visualization by multidimensional scaling (MDS) 
          of EMD distances of mean 2D-MHs as reported in Table \ref{tabAvgEMD}. 
          (a): Real and synthetic fingerprints. 
          The 2D-MHs originating from synthetic prints are labelled as D, H, L. 
          (b): Separate MDS visualization for the mean 2D-MHs of the real fingerprints only. 
          \label{MinutiaeAngleDist-hist-WassersteinMDS:fig}} 
\end{figure}

\subsection{Results}\label{Results:scn}

\label{secResults}

\begin{table}[ht!]
\begin{center}
\begin{tabular}{|c|c|c|c|c|c|c|c|c|} \hline
\multicolumn{9}{|l|}{Feature: 2D-MH} \\ \hline
& \multicolumn{2}{c|}{{DB 1 vs. DB 4}} & \multicolumn{2}{c|}{DB 2 vs. DB 4} & \multicolumn{2}{c|}{DB 3 vs. DB 4} 
& \multicolumn{2}{c|}{Average} \\ \hline
        & Set II & Set III & Set II & Set III & Set II & Set III & Set II & Set III\\ \hline
FVC2000 & 93.3  & 90.0  & 95.0  & 83.8  & 98.3  & 95.0 & 95.5 & 89.6\\ \hline
FVC2002 & 86.7  & 88.8  & 90.0  & 86.3  & 85.0  & 78.8 & 87.2 & 84.6\\ \hline
FVC2004 & 81.7  & 55.0  & 81.7  & 67.5  & 90.0  & 70.0 & 84.5 & 64.2\\ \hline \hline
\multicolumn{9}{|l|}{Feature: 2D-MH} \\ \hline
\multicolumn{7}{|l|}{}  & Set II & Set III\\ \hline
\multicolumn{7}{|l|}{FVC2000 DB1 vs. synthetic images according to Bicz \cite{Bicz2003}} & 83.3 & 92.5\\ \hline \hline
\multicolumn{9}{|l|}{Features: 2D-MH, interridge distances and percentage of bifurcations} \\ \hline
& \multicolumn{2}{c|}{{DB 1 vs. DB 4}} & \multicolumn{2}{c|}{DB 2 vs. DB 4} & \multicolumn{2}{c|}{DB 3 vs. DB 4} 
& \multicolumn{2}{c|}{Average} \\ \hline
FVC2000	& 100.0 & 92.5  & 100.0 & 87.5  & 100.0 & 97.5 & 100.0& 92.5\\ \hline
FVC2002 & 100.0 & 97.5  & 100.0 & 95.0  & 100.0 & 90.0 & 100.0& 94.2\\ \hline
FVC2004 & 95.0  & 72.5  & 90.0  & 83.8  & 98.3  & 97.5 & 94.4 & 83.6\\ \hline
\end{tabular}
\caption{Classification performance (correctly classified templates in percent).
         Distances between minutia histograms are measured using EMD.
         \label{tabPerformanceOverview}}
\end{center}
\end{table}

The results on all available FVC databases show that the proposed method by extended 2D-MHs
is able to separate real from synthetic prints with very high accuracy. 
On the training sets of all databases of FVC2000 and FVC2002, 
the classification performance of the combined feature set was 100\%,
and for the corresponding test sets, the performance was in the range from 87.5 to 97.5\%.
On the whole, the image quality in the databases of FVC2004 is clearly lower 
compared to the quality of images in previous competitions. 
Hence, it it is more challenging to avoid errors during the automatic
extraction of minutiae from theses images. We used a commercial off-the-shelf software 
for minutiae extraction.

The good discriminative power of 2D-MHs alone is not surprising upon inspection 
of a 2D visualization by multidimensional scaling 
(e.g. \cite[Chapter 14]{MardiaKentBibby1979}) of the mutual 2D-MH distances 
from Table \ref{tabAvgEMD} in Figure  \ref{MinutiaeAngleDist-hist-WassersteinMDS:fig}. 
In the left display, MHs from real fingers cluster in the middle of the left side while MHs 
from synthetic fingerprints come to lie in the upper right (for the year 2000) 
and closer to the center towards the bottom (for the years 2002 and 2004). 
In fact, we can see that the algorithms leading to minutiae formation have obviously undergone changes 
between the years 2000 and 2004, however, only moving the MHs moderately closer to 'realness'. 
In FVC2002, SFinGe version 2.51 was used and in FVC2004 SFinGe version 3.0.

Additionally, we tested the performance of 2D-MHs for discriminating 
between artificial fingerprint images generated 
by the software of Bicz~\cite{Bicz2003} and real fingerprint images.
To this end, we generated 110 synthetic fingerprints which were grouped 
into three sets as listed in Table \ref{tabSets}, and compared them against 
the 110 images of FVC2000 database 1. 
On set II, a classification performance of $83.3\%$ was achieved
and on set III, $92.5\%$ of the images were correctly classified.

\section{Suggestions on How to Improve the Generation of Synthetic Fingerprints}

\label{secImprovement}

Recall from Section \ref{secConstructionAndReconstruction} that the SFinGe model silently extends the well-standing biological hypothesis
of fingerprint pattern formation due to three converging ridge systems 
(for a brief discussion, c.f. \cite{KueckenNewell2005}), 
to a multitude a converging ridge systems starting at random locations; 
this process is the governing principle for minutiae formation.
Our work shows that this remarkable hypothesis can be tested 
and the tests show that the minutiae pattern formation appears to be of a more complex nature. 
It would be of interest to test fingerprint images 
generated by \cite{KueckenChampod2013}, \cite{ZhaoJainPaulterTaylor2012} and other reseachers
using extended minutiae histograms, once they become available.

One possibility to improve SFinGe is to modify
the fingerprint generation process in such a way 
that the resulting fingerprint images have the following properties:

\begin{itemize}
 \item MHs of synthetic prints should follow the distribution of MHs estimated from a database of real fingers.  
 \item More subtly also considering minutiae types, the relation of endings to bifurcations should
       resemble the relation and its distribution observed in real fingers.
 \item Similarly, the distribution of interridge distances should be similar to those 
       of real fingers acquired at the same resolution.
\end{itemize}

\paragraph{Two Sides of the Same Coin} \label{labelTwoSidesOfTheSameCoin}
An alternative option for the creation of synthetic fingerprints that pass the proposed 'test of realness'
is to consider synthetic fingerprint generation as a reconstruction task. 
Traditionally, the construction of artificial fingerprints is associated with an unknown outcome 
regarding the number, location, direction and type of minutiae,
whereas reconstruction aims at the generation of a fingerprint image 
which has best possible similarity in terms of minutiae properties to a given template.
Basically, we propose to focus on the generation of a 'realistic' minutiae template 
and ``all these other things above shall be added unto''
by the existing reconstruction methods.
Here is a possible outline of this approach:

\begin{itemize}
 \item First, a feasible foreground and orientation field is constructed, e.g. 
       using a global model like \cite{VizcayaGerhardt1996,HuckemannHotzMunk2008}
       which is able to incorporate the empirical distribution of 
       observed pattern types (Henry-Galton classes) and singular points \cite{CappelliMaltoni2009}. 
 \item Secondly, a realistic number of minutiae $n$ is drawn from the empirical data 
       of minutiae in fingerprints with the previously determined foreground size and pattern type.
 \item Thirdly, an inital minutiae template is obtained by choosing $n$ points e.g. randomly 
       on the foreground as minutiae locations and for each location, the minutiae direction is set to the
       local orientation $\theta$ or $\theta + 180\degree$ by chance.
 \item Fourthly, the 2D-MH of the minutiae template is computed and it is modified iteratively 
       until it passes the 'test of realness', i.e. the EMD between the current template 
       and the average 2D-MH of real fingerprints is below an acceptable threshold.
       Modification operations are the deletion and addition of minutiae and 
       flipping of the minutiae direction by 180\degree.
       The implementation for the computation of the EMD allows to analyze the flow of mass,
       so that the bins in the 2D-MH can be indentified which contribute above the ordinary to the total costs,
       and thus, the minutiae pairs that are 'the most unlikely' in comparison to the empirical distribution.
 \item Fifthly, minutiae types (ending and bifurcation) are assigned, 
       based on the empirical distribution in real finger patterns.
 \item Finally, the fingerprint image is reconstructed using e.g. Gabor filters or the AM-FM model.
       The interridge distances are checked for deviations from the empirical data obtained from real fingerprints
       and if required, the interridge distance image is adjusted and the reconstruction step is repeated.
 \item Optionally, noise can be simulated for copies of the constructed image, 
       and if desired, they can be rotated, translated and nonlinearly distorted.
\end{itemize}

\section{Identification}

\label{secIdentification}

\begin{figure}
 \begin{center}
   \includegraphics[width=0.45\textwidth]{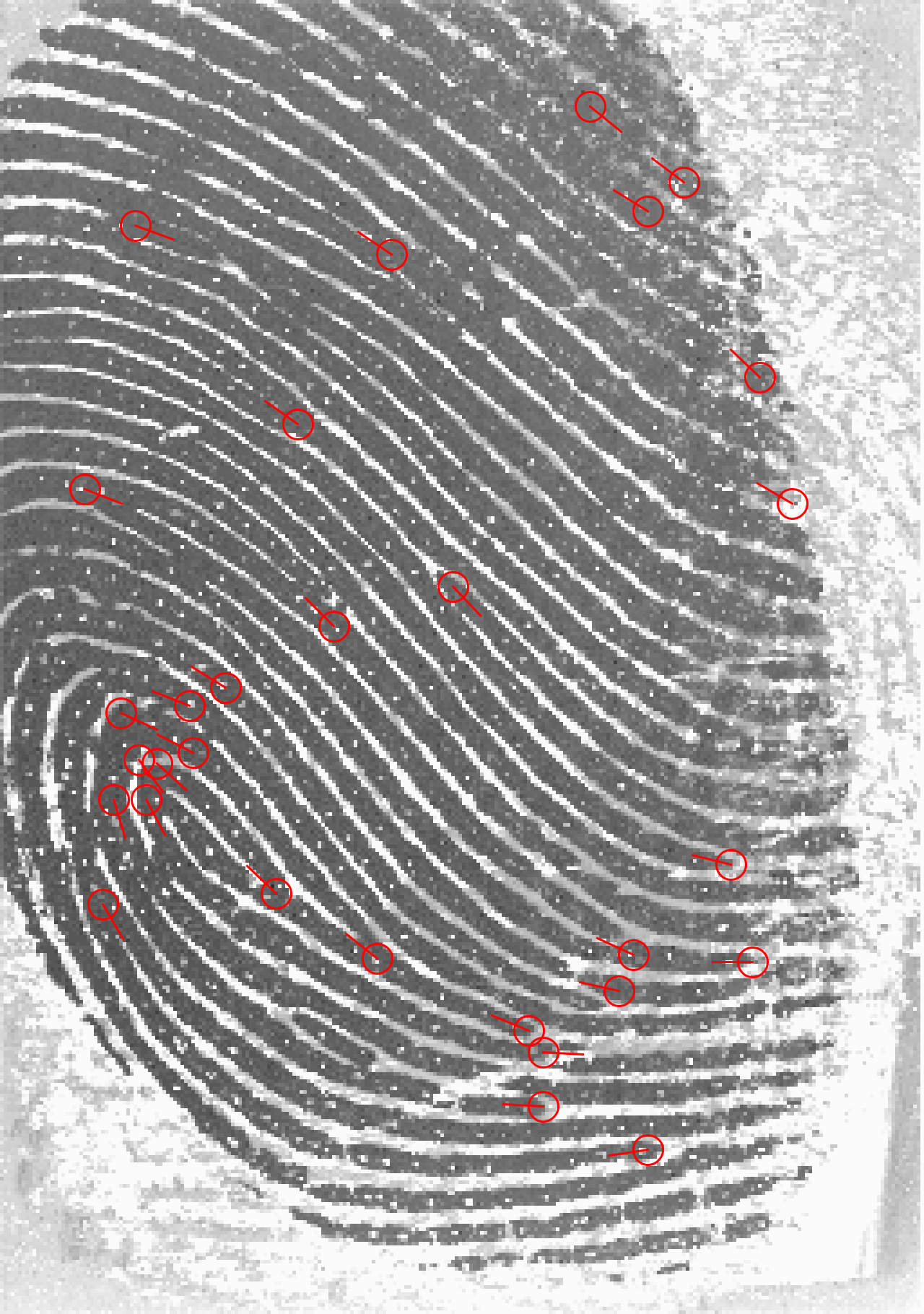} 
   \includegraphics[width=0.45\textwidth]{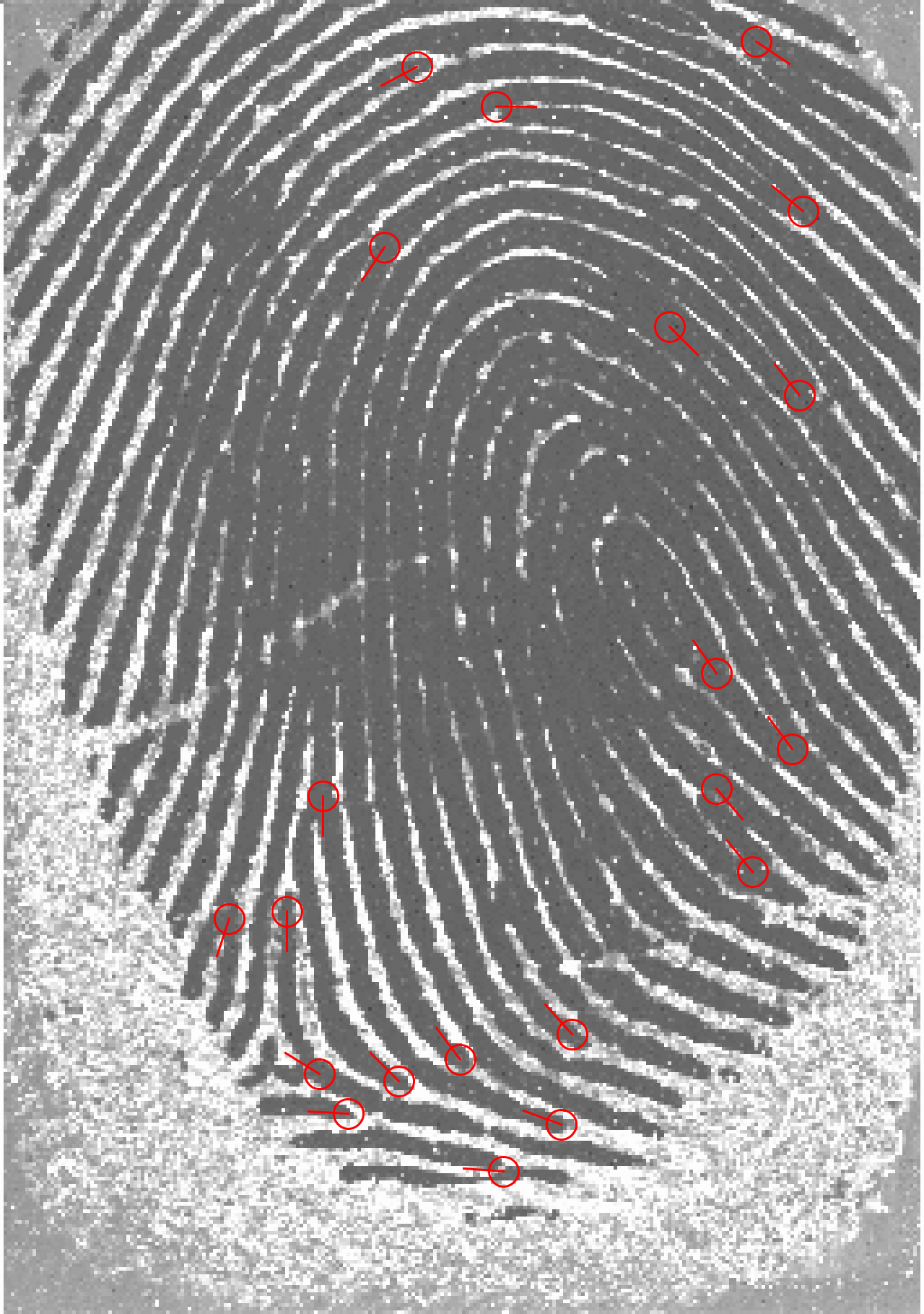}
 \end{center}
 \caption{In 597 out of 770 searches on FVC2000 DB2, the right finger ranked first.
          However, in a few cases a considerable part of the database had to be accessed in order to retrieve the template belonging to the query finger. In this example, the search with impression four (left) of finger 105 resulted in a small intersection with the minutia histogram of same finger (impression one, right). Missing and spurious minutiae as well as a small overlap between the aligned templates deteriorate the performance for the displayed query.
         \label{figIndex}}
\end{figure}

Up to this point, we applied minutiae statistics for classifying a fingerprint as real or synthetic.
In this section, we explore the potential of MHs for identifying individuals. 

The approach described previously based on MHs allows 
to represent a fingerprint as a fixed-length feature vector. 
Obtaining a fixed-length feature vector representation 
marks the grand goal for biometric template protection schemes
and has previously been achieved by Xu \textit{et al.} 
in their spectral minutiae representation \cite{XuVeldhuisBazenKevenaarAkkermansGokberk2009}. 
Robust fixed-length feature vectors also appear highly promising for an identification scenario 
in which a large database is searched for a query fingerprint.

Up to this point, the minutiae distributions represented by MHs were compared using
the earth mover's distance. For the purpose of identification, 
we compute the histogram intersection to better handle partial overlap \cite{SwainBallard1991}. 
This approach is related to ``geometric hashing'' \cite{LamdanWolfson1988}: 
a hash table is built by quantizing geometric objects like 
minutiae triplets which were used in \cite{GermainCalifanoColville1997,BhanuTan2003,LiangBishnuAsano2007}.

In a first test on FVC2000 DB2, we achieved 
average access rates (average part of the database that is accessed 
using an incremental search strategy until the corresponding finger is found) 
between 2.33\% and 4.01\% using unnormalized 4D-MHs 
with 20 bins for differences between minutiae directions, 
20 bins for Euclidean distances between minutiae locations, 
20 bins for the angle of the relative location of the second minutiae with respect to the first 
and 4 bins for minutiae type combinations. We chose this database, 
because it was used in tests measuring the indexing performance by other researchers: 
De Boer \textit{et al.} tested three different features (orientation field, finger code, minutiae triplets) 
and their combination on this database and they reported rates between 1.34\% and 7.27\% \cite{DeboerBazenGerez2001}. 
Cappelli \textit{et al.} proposed minutiae cylinder code 
and locality-sensitive hashing for fingerprint indexing \cite{CappelliFerraraMaltoni2011} 
and report an average access rate of 1.72\% for this database. 
As we used, in contrast to these two studies, an ``off the shelf'' minutiae extractor (\nolinebreak{VeriFinger~5.0}), 
and as the performance hinges on correct minutiae extraction (see Figure \ref{figIndex} and the discussion below) 
we deem our results very promising upon more ``correct'' minutiae extraction.  

In our test, we used the sum of intersections 
between corresponding bins as a score (BIS = bin intersection score)
and sorted the list of fingers in the database in descending order. 
In 597 out of 770 searches (77 $\%$), the correct finger was ranked first in the list. 
If the search is narrowed down to templates with 30 or more minutiae, 
than it ranked first in 393 out of 441 searches (89 $\%$). 

We inspected the cases in which a larger portion 
of the database were accessed. 
It turned out that the main reason lies in minutiae extraction errors (missing and spurious minutiae) which, 
of course, have a negative impact on the score: missing minutiae reduce the intersection 
between minutiae histograms of templates from the same finger,
spurious minutiae can increase the intersection between histograms from templates of different fingers.   
A typical example is shown in Figure \ref{figIndex}. 
Another reason lies in almost no overlap areas between the two prints compared. 
We stress at this point that our BIS has been designed to alleviate small overlaps. 

It is obvious that for a fair comparison between different minutiae-based indexing methods, 
the same minutiae templates have to be used. 
In doing so, the influence of different minutiae extractors 
on the identification performance can be eliminated and only then results become comparable.
It is of interest to quantify the impact of different minutiae extractors 
and different fingerprint image enhancement techniques on the identification performance,
but these comparisons are beyond the scope of this study.
This first test shows the suitability of MHs for identification purposes
and this direction deserves further research.

\section{Fingerprint Individuality and Quantifying Weight of Evidence in Court}

\label{secIndividuality}

Universality, collectability, permanence and uniqueness are properties which render an anatomical or behavioral trait useful for biometric recognition~\cite{HandbookFingerprintRecognition2009}. 
Permanence of the fingerprint pattern was scrutinized by Galton~\cite{Galton1892} 
more than a century ago and it was later confirmed that the pattern's development is finalized at an estimated gestational age of 24 weeks~\cite{Babler1991}.
Uniqueness of fingerprints is commonly assumed by all researchers and practitioners dealing with fingerprint recognition.
Fingerprint individuality has never been proven, but there is a long history
of models attempting to explain and quantify fingerprint individuality 
starting with Galton in 1892 \cite{Galton1892} to the present day. 
Stoney gives an overview over 10 major models formulated between 1892 and 1999
in Chapter 9 of \cite{LeeGaensslen2001}. For recent additions, 
please see \cite{PankantiPrabhakarJain2002,NagarChoiJain2012,KueckenChampod2013} and the references therein.
There are two broad categories of models: 
First, mathematical models trying to encompass the distribution of features 
extracted from observed prints.
And secondly, biology based models about randomness during the 
formation of friction ridge skin in prenatal development 
of human life \cite{WertheimMaceo2002,KueckenNewell2004,KueckenNewell2005,Kuecken2007,KueckenChampod2013}.
Notwithstanding the lack of 'proof' of uniqueness, fingerprints have a long success story 
in commercial, governmental and forensic applications \cite{FingerprintSourcebook2011}.

As a consequence of an on-going reformation process \cite{StrengtheningForensicScience2009}, 
in the future forensic experts may have to quantify the weight of evidence 
with probabilities and errors rates instead of a binary decision \cite{NeumannEvettSkerrett2012}.
By their very nature, MHs (2D-, 4D-, second and higher order) as empirical realisations of an underlying -- to date unknown -- true minutiae distribution precisely yield such probabilities. For instance for a given confidence level $0<\alpha<1$, for each true finger $t$ multiply represented in a database, a neighborhood $N_{\alpha,t}$ in the space of corresponding (possibly higher order and/or extended) MHs can be computed, via bootstrapping, say, such that for a query fingerprint $q$ with MH $h_q$ under the null-hypothesis that $q$ is another imprint of $t$, 
$$ \mathbb P_e\{h_q\in N_{\alpha,t}\}\geq 1-\alpha$$
where $\mathbb P_e$ is the empirical probability.

The use of the corresponding test, of course, relies on prior thorough statistical studies of these true distributions. Further research, exploiting the abundance of optional statistical methods, may choose likelihood ratios \cite{NeumannEtAlii2007}, say, for improved quantification.

A crucial point is that the underlying minutiae statistics have to be based on a large number of real fingers. 
Minutiae should be manually marked or in semi-automatic fashion, a human should 
inspect automatically extracted minutiae in order to avoid minutiae extraction errors
and their influence on the MHs.
Ideally, interridge distances or related measures like e.g. ridge count would be taken into account
and additional information would be available for minutiae templates,
including age, body height, sex and ethnicity \cite{GutierrezrredomeroEtAlMinutiaeFrequencies2012}.
This would enable the computation of more sophisticated statistics, 
e.g. the probability that a print with certain minutiae histogram stems from a person of certain age group. 

In summary, MHs can become a useful tool for forensic experts 
who are requested to quantify the weight of fingerprint evidence in court
based on empirical ground truth data.

\section{Conclusion}
\label{Discussion:scn}

The proposed MHs capture and comprise 
relevant information of fingerprints, namely the full distribution of minutiae, which enable to separate 
current state-of-the-art synthetic fingerprints from prints of real fingers.
This study reveals a fundamental difference between natural finger pattern formation
and state-of-the-art synthetic fingerprint generation processes.
As a consequence, any results obtained on existing databases of synthetic fingerprints 
should be regarded with caution 
and may not reflect the performance of a fingerprint comparison software in a real-life scenario.
A performance evalution of FVC2004 showed that 'the behavior of the algorithms over
DB4' (the database consisting of synthetic prints)  
'was, in general, comparable to that on the real databases' \cite{CappelliMaioMaltoniWaymanJain2006}.
However, this can also be interpreted as an indicator that the participating algorithms
are not yet optimized for the specifics of the empirical distribution of minutiae in real fingerprints.     

\subsection{Other Research Directions}

Another direction of research beyond the ones discussed 
in Sections \ref{secImprovement} to \ref{secIndividuality}
can go towards the identification of additional features 
for the discrimination between real and synthetic prints. 
In the context of presentation attack detection, 
it would be highly desirable to detect a reconstructed fingerprint,
even if this property cannot be infered from the minutiae distribution.

We would also like to explore the suitability of minutiae histograms 
for security applications, e.g. in combination with the fuzzy commitment scheme or
the fuzzy vault scheme, and for the generation of cancelable templates.

In this research, we used a second order minutiae statistic 
reflecting the covariance structure of the minutiae distributions 
which turned out to be highly discriminatory for our task. 
Exploring the discriminative potential of higher order minutiae histograms is beyond the scope of this paper 
and remains an interesting endeavor.

\subsection{Acknowledgements}

The authors thank Raffaele Cappelli and Davide Maltoni for sharing the images shown in Figure \ref{figSurvey},
Pedro Vizcaya for sharing the images displayed in Figure \ref{figSynthetic} (a) and (b), 
and Wieslaw Bicz for sharing the software to generate the images in Figure \ref{figSynthetic} (c) to (f).
The authors gratefully acknowledge the support of the 
Felix-Bernstein-Institute for Mathematical Statistics in the Biosciences
and the Volkswagen Foundation.

\end{document}